\documentclass[sigconf]{acmart}

\usepackage{graphicx}
\usepackage{float}
\usepackage{amsmath}
\usepackage{multirow}
\usepackage{tablefootnote}
\usepackage{booktabs}
\usepackage{threeparttable}
\usepackage{subfig}
\usepackage[export]{adjustbox}
\usepackage{balance}
\graphicspath{{tex_folder/figures/}}
\usepackage{pifont}
\newcommand{\cmark}{\ding{51}}%

\AtBeginDocument{%
  \providecommand\BibTeX{{%
    \normalfont B\kern-0.5em{\scshape i\kern-0.25em b}\kern-0.8em\TeX}}}

\copyrightyear{2022}
\acmYear{2022}
\setcopyright{acmcopyright}
\acmConference[MM '22] {Proceedings of the 30th ACM International Conference on Multimedia }{October 10--14, 2022}{Lisboa, Portugal.}
\acmBooktitle{Proceedings of the 30th ACM International Conference on Multimedia (MM '22), October 10--14, 2022, Lisboa, Portugal}
\acmPrice{15.00}
\acmISBN{978-1-4503-9203-7/22/10}
\acmDOI{10.1145/3503161.3548312}

\begin{document}

\title{Digging Into Normal Incorporated Stereo Matching}

\author{Zihua Liu}
\authornote{Both authors contributed equally to this work.}
\authornotemark[0]

\affiliation{%
  \institution{Tokyo Institute of Technology}
  \city{Tokyo}
  \country{Japan}
}
\email{zliu@ok.sc.e.titech.ac.jp}

\author{Songyan Zhang}
\authornotemark[1]

\affiliation{%
  \institution{CAD Research Center, Tongji University.}
  \city{Shanghai}
  \country{China}}
\email{spyder@tongji.edu.cn}

\author{Zhicheng Wang}
\affiliation{%
  \institution{CAD Research Center, Tongji University.}
  \city{Shanghai}
  \country{China}}
\email{zhichengwang@tongji.edu.cn}

\author{Masatoshi Okutomi}
\authornote{Corresponding author.}
\authornotemark[0]

\affiliation{%
  \institution{Tokyo Institute of Technology}
  \city{Tokyo}
  \country{Japan}
}
\email{mxo@ctrl.titech.ac.jp}

\renewcommand{\shortauthors}{Zihua Liu, Songyan Zhang, Zhicheng Wang, \& Masatoshi Okutomi}

\begin{abstract}
Despite the remarkable progress facilitated by learning-based stereo-matching algorithms, disparity estimation in low-texture, occluded, and bordered regions still remains a bottleneck that limit the performance. To tackle these challenges, geometric guidance like plane information is necessary as it provides intuitive guidance about disparity consistency and affinity similarity. In this paper, we propose a normal incorporated joint learning framework consisting of two specific modules named non-local disparity propagation(NDP) and affinity-aware residual learning(ARL). The estimated normal map is first utilized for calculating a non-local affinity matrix and a non-local offset to perform spatial propagation at the disparity level. To enhance geometric consistency, especially in low-texture regions, the estimated normal map is then leveraged to calculate a local affinity matrix, providing the residual learning with information about where the correction should refer and thus improving the residual learning efficiency. Extensive experiments on several public datasets including Scene Flow, KITTI 2015, and Middlebury 2014 validate the effectiveness of our proposed method. By the time we finished this work, our approach ranked 1st for stereo matching across foreground pixels on the KITTI 2015 dataset and 3rd on the Scene Flow dataset among all the published works.

\end{abstract}



\begin{teaserfigure}
  \includegraphics[width=\textwidth]{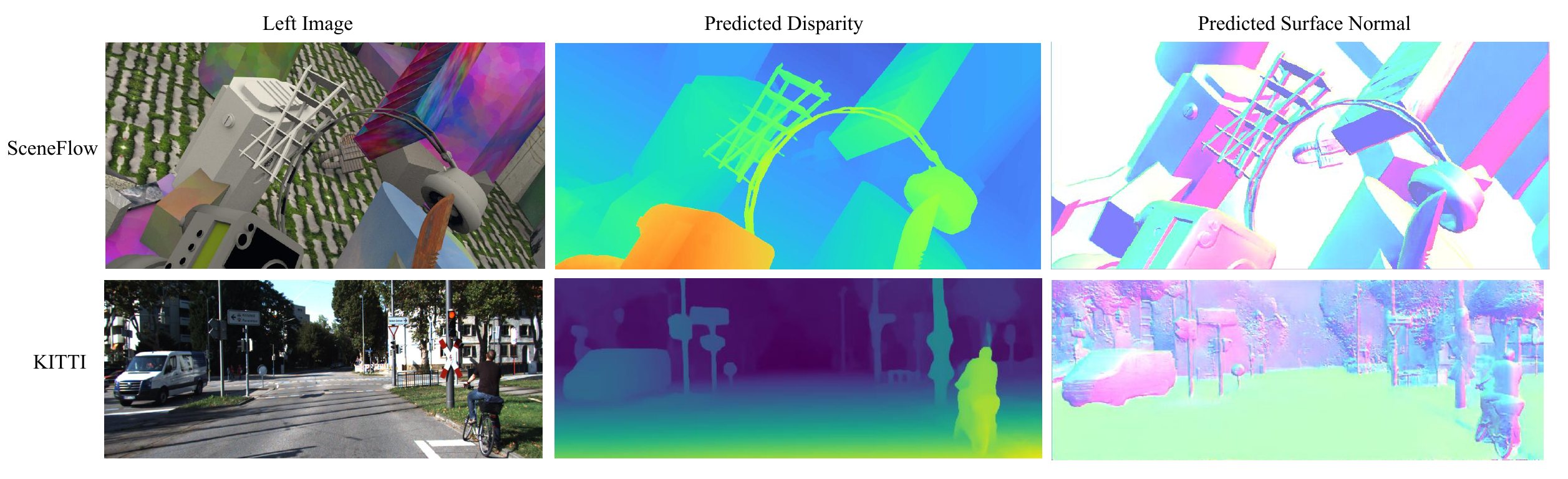}
  \caption{Visualization of predicted disparity and surface normal on SceneFlow and KITTI 2015 datasets.Our proposed NINet successfully preserves detailed structure and consistency even in the challenging bordered, low-texture and occluded regions.}
  \Description{}
  \label{fig:teaser}
\end{teaserfigure}

\maketitle


\section{Introduction}
Stereo matching has been a fundamental task in computer vision for decades of years. Given a pair of rectified stereo images, the core of stereo matching is to find the corresponding pixels along the epipolar line so that the disparity can be calculated, which has been closely related to various applications like robotic navigation, autonomous driving, and VR/AR. 

Recently, the application of convolutional neural networks significantly facilitates the development of stereo matching \cite{psmnet, gcnet, AANET, EDNet, dispnetc}. Most of the learning-based methods can be divided into two categories according to the construction of cost volume which is an essential step for stereo matching as mentioned in \cite{taxonomy2001}. The pioneer works \cite{dispnetc, flownet} employ the inner production for measuring the similarity between pixels and propose a lightweight correlation layer for building the cost volume which is widely adopted in the following works \cite{AANET, fadnet, consistency}. The correlation layer takes advantage of efficient 2D convolution networks but suffers from the loss of context information is inevitable as only a single channel of feature is preserved. GCNet proposes to build a 4D cost volume where the complete stereo features are directly concatenated along the disparity dimension for abundant context information. Most of the top-performance methods like \cite{gwcnet, ganet, acvnet} leverage this concatenation-based cost volume which follows 3D convolution networks to improve stereo matching. Even though that learning-based methods help the disparity estimation reach a new peak, ambiguous matching still exists especially in bordered, occluded, and low-texture regions, which remains a challenge. EDNet\cite{EDNet} proposes a novel method to learn an error-aware attention map which is utilized to guide the residual learning concentrate on the inaccurate regions. However, there still exists a limitation that the model only has intuitive information about where the error occurs but lacks guidance about where the correction should be learned from. Some previous works manage to improve the disparity estimation with stronger geometric constraints \cite{geonet, normal_assisted, VNL} by conducting a joint learning framework with the surface normal, but most of them just simply enforce the consistency between normal estimation and disparity estimation. The normal regression is usually independent of the stereo matching which means the plane information fails to be fully utilized by stereo matching.

To tackle the challenge of stereo matching in bordered, occluded, and low-texture regions as well as explicitly leverage normal information, in this paper, we propose a normal incorporated joint learning network named NINet. Our idea lies that the relatively easier task of normal estimation can provide abundant plane information as well as pattern affinity. And disparities within the same plane should be consistent. To realize this, we propose two specific modules which are the non-local disparity propagation(NDP) module and affinity-aware residual learning(ARL) module, respectively. The former NDP module leverages the estimated normal and computes a non-local affinity matrix and a corresponding location offset indicating the sampled points for spatial propagation at the disparity level. This operation can efficiently improve the disparity estimation performance, especially in bordered and occluded regions. Moreover, to mitigate the effect of low-texture regions, we provide another ARL module to further correct the disparity estimation. Following EDNet \cite{EDNet}, we introduce error based attention method with the modification of utilizing the normal estimation to calculate a local affinity matrix. Using the reweighted feature map based on the local affinity matrix to regress the disparity residual.
To sum up, our contributions can be summarized as follows:

\begin{itemize}
    \item We propose a normal incorporated joint learning framework to explicitly leverage normal estimation and exploit plane information to provide intuitive geometric guidance for disparity estimation.
    \item We propose two specific modules: non-local spatial propagation (NDP) at the disparity level and affinity-aware residual learning (ARL) at the feature level. The former module efficiently improves the disparity estimation in both bordered and occluded regions while the latter module primarily improves the performance in low-texture regions.
    \item We propose a novel method to generate dense pseudo normal ground truth on sparsely annotated KITTI 2015 datasets which enables the application of our method in real-world scenes.
    \item We conduct extensive experiments on several public datasets including Scene Flow, KITTI 2015, and Middlebury 2014. Our proposed NINet outperforms most of the previous top-performance approaches and ranks 1st on KITTI 2015 leaderboard for foreground pixels. Our method demonstrates remarkable generalizability as well, which further validates the effectiveness of our method.
\end{itemize}


\begin{figure*}
	\centering
	\includegraphics[width=0.93\linewidth]{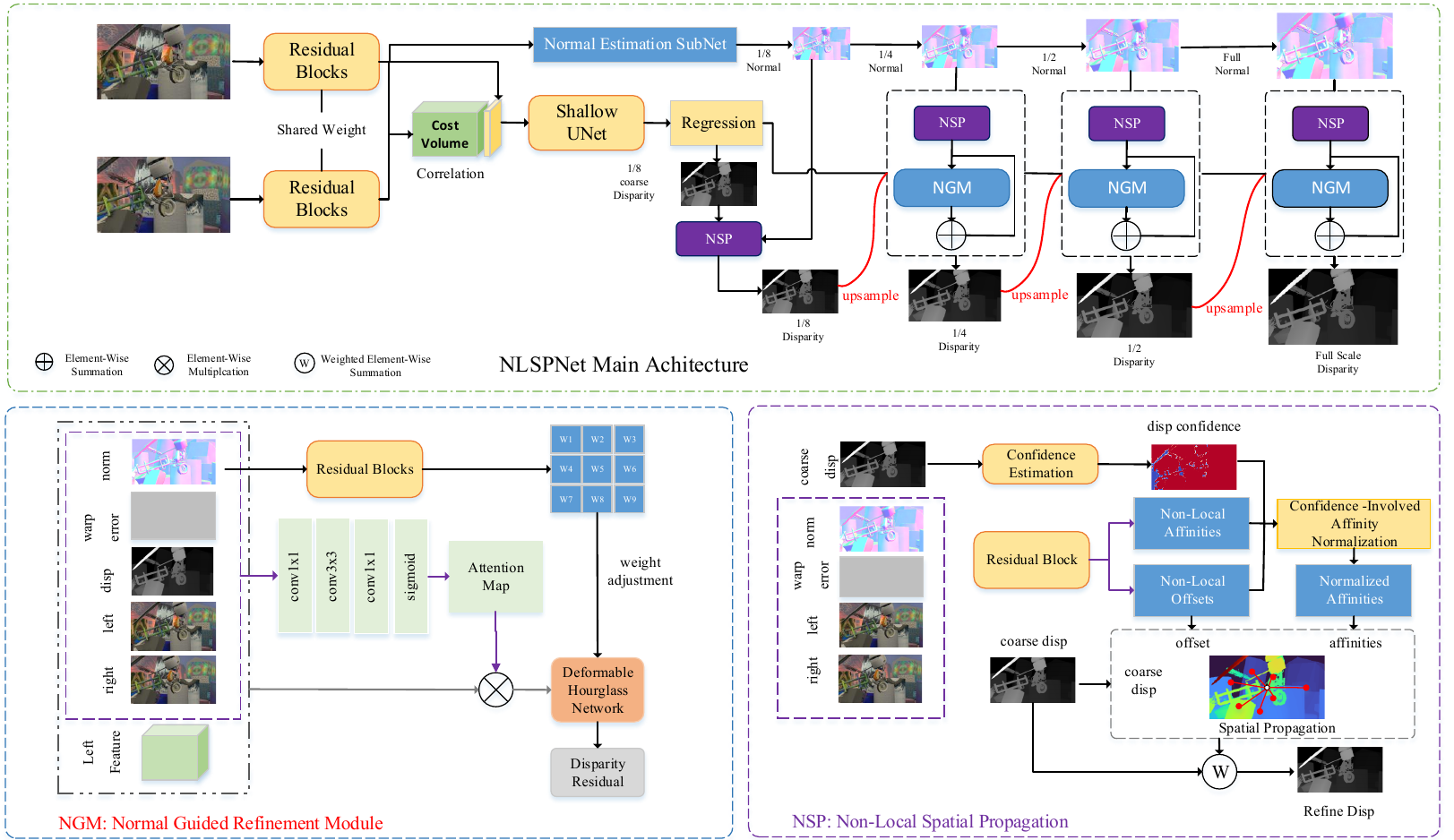}
	\caption{An overview of our proposed NINet. Our model is mainly composed of two modules named ARL and NDP. Note that some skip-connection operations are omitted here for simplifying the visualization.}
	\label{fig:architecture}
	\vspace{-0.8 em}
\end{figure*}
\section{Related Works} \label{sec:related_work}
 In this section, we briefly review some pioneer learning-based works for disparity estimation as well as giving an introduction about surface normal incorporated stereo matching methods. Finally, we summarize the spatial propagation mechanism utilized in this field. 
\subsection{Learning-based Stereo Matching}
The pioneer learning-based work DispNetC\cite{dispnetc} follows the previous work \cite{flownet} which introduces a correlation layer to calculate the inner product of the corresponding pixels for measuring the similarity. Many real-time approaches like \cite{fadnet, AANET} takes this method for constructing the cost volume. To preserve abundant context information which may be lost in the correlation operation, GCNet \cite{gcnet} employs the concatenation of the left and right features along the disparity dimension to construct a 4D cost volume. The concatenation-based cost volume following 3D convolution networks for aggregation are widely adopted in many state-of-the-art works like \cite{ganet, CSPN, CFNet, cascade_cost_volume}. PSMNet\cite{psmnet} proposes a spatial pyramid pooling module for incorporating global context information into image features along with stacked hourglass 3D convolution networks for extending the context information. As 3D convolution networks are highly memory-consuming, StereoNet \cite{stereonet} decreases the layer number of 3D convolution networks and improves the performance by using RGB images for edge refinement. To incorporate both the advantages of correlation-based cost volume and concatenation-based volume, GwcNet\cite{gwcnet} proposes the group-wise correlation volume to form a 4D correlation-based cost volume which is then concatenated with the concatenation cost volume. The recent remarkable work ACVNet\cite{acvnet} introduces an attention-based strategy to suppress redundant information in the cost volume which achieves top performance and real-time efficiency. 
\subsection{Normal Assisted Depth Estimation}
Many indoor depth estimation utilizes the normal as an assisted geometry constraints \cite{pattern_affinity, normal_assisted, deep_for_normal, geonet, graph_normal}.  GeoNet \cite{geonet} proposes a normal-to-depth branch and a depth-to-normal branch for jointly learning. VNL \cite{VNL} designs a loss term named virtual normal directions to impose geometric constraints. The work \cite{normal_assisted} is also a joint learning-based model which proposes a consistency loss to refine the depth from depth/normal pairs. An energy optimization-based approach is introduced in \cite{energy_minimizing} for multi-view stereo matching. IDNSolver \cite{energy_minimizing} proposes an iterative energy optimization strategy which is a novel view for multi-view stereo matching. The ASNDepth \cite{ASNDepth} introduces an adaptive surface normal constraint to effectively correlate the depth estimation with geometric consistency. Most related works follow the pipeline of enforcing the consistency between the depth and the normal to impose more implicit geometric constraints. 

\subsection{Spatial Propagation for Disparity}

Spatial propagation is a popular mechanism in depth completion whose input data is a sparse depth map generated by the sensor of LiDAR. Representative works are \cite{CSPN, NLSPN, DSPN}. When it comes to the depth estimation task, PAPNet\cite{pattern_affinity} conducts two types of propagation which are task-specific propagation and cross-task propagation, respectively. The cross-task propagation integrates affinity from different tasks including semantic segmentation, surface normal estimation, and depth estimation while the task-specific propagation performs an iterative diffusion in the feature space to widely spread the cross-task affinity patterns. The recent work PatchMatchNet \cite{patchmatchnet} introduces an iterative multi-scale patchmatch algorithm to learn adaptive propagation for multi-view stereo matching. HitNet \cite{HitNet} imitates its predecessor \cite{PM} and performs spatial propagation with the assistance of predicted slanted plane hypotheses.

\section{Methodology} \label{sec: method}
In this section, we will provide a thorough introduction to our proposed normal incorporated stereo matching network(NINet) in the following order: the architecture of our model is demonstrated in Subsection \ref{sub: network} following the instruction of normal estimation subnet in Subsection \ref{subsec:normal}. Details about our proposed two modules can be found in Subsection \ref{subsec:NDP} and Subsection \ref{subsec: ARL}. We describe our multi-scale training mechanism in Subsection\ref{subsec:multiloss}.

\begin{figure*}
    \centering
    \includegraphics[width=0.98\linewidth]{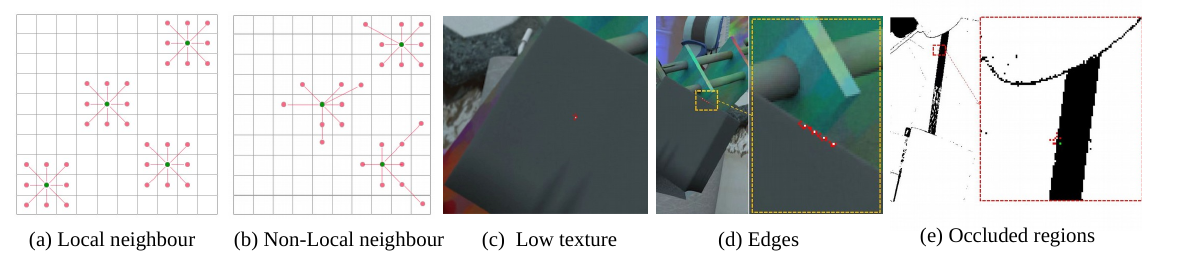}
    \caption{Illustration of local spatial propagation (a), non-local spatial propagation (b). (c) shows sampled points in low-texture regions. (d) demonstrates sampled points at edges. (e) shows sampled points in occluded regions. Red points indicated the selected ones for propagating disparity to the targeted white/green point. It's obvious that our method successfully learns to dynamically sample points for propagation according to different patterns.}
    \label{fig:ndp}
\end{figure*}

\subsection{Network Architecture} \label{sub: network}
The overview architecture of our proposed NINet is shown in Figure \ref{fig:architecture}. We proposed a normal incorporated joint learning framework for surface normal and disparity estimation. A weight-sharing 2D residual blocks are used for feature extraction from stereo images which are then fed into two parallel branches for normal estimation and disparity estimation, respectively. The normal estimation will be discussed in Subsection \ref{subsec:normal}. As for the stereo matching process, we imitate \cite{fadnet, dispnetc} and employ inner production for calculating the similarity between pixels. Given a pair of stereo feature maps, $i.e.$ $f_l$ and $f_r$ $\in R^{H\times W \times C}$, the correlation is computed as :
\begin{align}
    Coor(d, x, y) = \frac{1}{C}<f_l(x-d,y), f_r(x,y)>,
\end{align}
where $<A,B>$ indicates the inner production between $A$ and $B$. The aggregation part takes a shallow UNet-like network which follows our proposed NDP and ARL modules for disparity estimation at multi scales. 

\subsection{Normal Estimation Sub-Net} \label{subsec:normal}
Since there is a certain degree of the geometric relationship between disparity and surface normal, we directly use the network structure of DispNetC\cite{dispnetc} to estimate the surface normal and use the coarse-to-fine strategy to obtain the surface normal prediction results at different scales with a normalization at the feature dimension. For the reason that Sceneflow and KITTI datasets do not provide ground truth for surface normal, we use a Sobel operator to obtain plane information as the supervision information. See Section \ref{subsec:nes} for the different processing methods of different datasets.

\subsection{Non-Local Disparity Propagation} \label{subsec:NDP}
Inspired by the spatial propagation mechanism utilized in the depth completion task[4], we introduce a normal incorporated non-local disparity propagation module in which we hub NDP to generate non-local affinities and offsets for spatial propagation at the disparity level. The motivation lies that the sampled pixels for edges and occluded regions are supposed to be selected. The propagation process aggregates disparities via plane affinity relations, which alleviates the phenomenon of parallax blurring at object edges due to frontal parallel windows. And the disparities in occluded areas are also be optimized at the same time by being propagated from non-occluded areas where the predicted disparities with high confidence. The architecture of NDP module can be seen in Figure \ref{fig:architecture}.

In the NDP module, the coarse disparity and its corresponding warped error are firstly utilized to obtain the disparity confidence map through the confidence estimation branch. A sigmoid activation function is used to generate the disparity confidence map. This confidence map can reflect the reliability of the current disparity which benefits the spatial propagation process.

For the disparity spatial propagation, previous works such as \cite{CSPN} use a local fixed $3\times3$ window to consider all the possible directions of neighbor pixels for disparity aggregation. Its process can be shown in Figure3.a. For a specific location $p$, we have its 8-closest neighbor $N_{p}$:
\begin{align}
    N_{p} =\left\{p+p_n |p_n\in\left\{(-1,-1),(-1,0)...(1,0),(1,1)\right\}\right\} 
\end{align}

After we obtain the affinity map $A\in R^{H\times W \times 8}$ and confidence map $C\in R^{H\times W \times 1}$. The local spatial propagation for a specific position $p$ can be described as follows:
\begin{align}
    \tilde{d}_{p} = \sum_{n\in N_{p}} (A_{p}(n)\times c_n \times d_n)+(1-\sum_{n\in N_{p}}(A_{p}(n)\times c_n)\times d_p
    \label{eq:eq2})
\end{align}
where $n$ enumerates the locations in $N_{p}$. $A_{p}(n)$, $c_n$ is the corresponding affinity and confidence. $d_p$ is the initial disparity, and $\tilde{d_p}$ is the refined disparity.

Unlike conventional spatial propagation using fixed propagation window, in NDP, we directly use the concatenation of pixel-level information to provide the guidance to learn an additional offset $\Delta p$ and its affinities for each 8 closest neighbors. By applying the non-local offset, the propagation kernel can break through the limitation of the fixed window, and thus enhances the ability to determine the position of the propagation neighborhood adaptively which can be seen in Figure3.b. The new 8 neighbors can be defined as:
\begin{align}
    \tilde{N}_{p} = \{n+\Delta p_n | n\in N_{p}, \Delta p_n=f_{\theta}(I_l, I_r, S, E)\}
\end{align}
Here $\tilde{N}_{p}$ is the non-local 8 closest neighbour, $f$ is the residual block in Figure \ref{fig:ndp} with a learnable parameter $\theta$. The network takes left image $I_l$, right image $I_r$, estimated surface normal $S$, and warped error $E$ as the input to provide the guidance information. Here the surface normal indicates the geometry consistency relations, and the warped error $E$ reveals the locations of the inaccurate and occluded areas, which is beneficial to increase the directivity and efficiency of the spatial propagation process. Besides, the residual block also predicts the non-local affinities to manifest the similarities between non-local neighbor's disparities and the target disparity. To enhance the stability of the propagation, we apply confidence incorporated affinity normalization by multiplying the predicted non-local affinity values with their corresponding confidence value and then be divided by the absolute sum of all 8 non-local affinities at the target position. For one specific location $p$, the normalized affinity $\tilde{A}_{p}^{norm}$ is:
\begin{align}
    \tilde{A}_{p}^{norm} = \left\{\frac{a_ic_i}{\sum_{i=1}^8|a_ic_i|}|a_i\in \tilde{A}_p, c_i \in C \right\}
    \label{eq:a_norm}
\end{align} 
By using the non-local neighbour $\tilde{N}$ and normalized non-local affinity $\tilde{A}_{p}^{norm}$, we apply non-local disparity propagation to refine the disparities. Compared with Equation \ref{eq:eq2} ,the non-local propagation process becomes:
\begin{align}
    \tilde{d}_p =\sum_{n\in\tilde{N}_{p}}(\tilde{A}_{p}^{norm}(n)\times d_n)+(1-\sum_{n\in\tilde{N}_{p}}\tilde{A}_{p}^{norm}(n))\times d_{p}
\end{align}
As it shows that we also assign a dynamic weight to the initial disparity which equals 1 minus the summation of the affinities of its 8 non-local neighboring points.

Figure \ref{fig:ndp} (c)(d)(e) show how our proposed NDP module decides non-local neighbor of specific positions when dealing with edges, low texture, and occluded regions with the guidance of surface normal and warped error. Thus, the proposed NDP module can refine the disparities in these ill positions by non-local propagation.

\subsection{Affinity-Aware Residual Learning} \label{subsec: ARL}
The non-local disparity propagation module primarily concerns the disparity estimation in bordered and occluded regions, but stereo matching in low-texture regions remains a bottleneck. To mitigate this problem, we imitate the idea from EDNet\cite{EDNet} to apply error based spatial attention strategy for efficient residual learning and obtain the error concentrated feature map $\hat{F_e} \in R^{H\times W \times C}$. Given the concatenation of stereo images, warped error predicted disparity, and surface normal map, a block of 2D convolution network is introduced to obtain an error-aware feature map $F_e$. A sigmoid function is then used for calculating an error-based spatial attention map. The error concentrated feature map $\hat F_e$ is computed as:
\begin{align}
    \hat{F_e} = F \odot (\sigma{F_e}),
\end{align}
where $F$ is the original feature map, $\odot$ represents element-wise product, and $\sigma()$ is sigmoid function. 

However, it's insufficient to know where mismatching occurs as the model still lacks intuitive guidance about where to refer for better refinement. Therefore, with the concatenation of stereo images, predicted disparity, warped error map, and normal estimation, we conduct a residual block to calculate a normalized local affinity matrix $A\in R^{H\times W \times k\times k}$ within a $k\times k$ neighboring region for each pixel. The error-concentrated feature map is further re-weighted with this affinity matrix. The final local affinity-aware feature map is calculated as follows:
\begin{align}
    \tilde{F}_{i,j} = \sum_{a,b=-(k-1)/2}^{(k-1)/2} \hat{F}_{i+a,j+b} \odot A_{i,j}(a,b),
\end{align}
where $\odot$ represents the element-wise product. Assisted by this normal guided local affinity matrix, those mismatched pixels in low-texture regions can be corrected by referring to their neighboring regions with similar affinity. The re-weighted feature map is then fed into a deformable hourglass network for further aggregation and get an affinity-aware disparity $r_a$ to obtain the updated disparity estimation.

\subsection{Multi-Scale Loss Function} \label{subsec:multiloss}
Our network is trained in an end-to-end manner supervised by ground truth disparity and generated ground truth surface normal at multi scales. Disparity loss $L_s^d$ and surface normal loss $L_s^n$ at scale $s$ are computed as:
\begin{align}
 L_s^d =\frac{1}{N}\sum_{i=1}^N
 {SmoothL1(d_i^s-\hat{d_i^s})}
\end{align}
\begin{align}
 L_s^n =\frac{1}{N}\sum_{i=1}^N
 {SmoothL1(n_i^s-\hat{n_i^s})}
\end{align}
where $N$ is the number of elements for the predicted map. $\hat{d_i^s}$ and $\hat{n_i^s}$  indicate the $i^{th}$ element of the estimated disparity $\hat{d^s}$ and estimated surface normal $\hat{n^s}$ while $d^s$ and $n^s$ represent the ground truth for disparity and surface normal, respectively. The utilized Smooth L1 loss is defined as follows:
\begin{align}
SmoothL1(x)=\begin{cases}
0.5x^2, &if \lvert x \rvert<1\\
 \lvert x \rvert -0.5, &otherwise
\end{cases}
\end{align}

To supervise the confidence $c$ used in the NDP module, we apply the loss function in \cite{HitNet}:
\begin{align}
 L_s^c(c) = max(1-c,0)_{\tau\lvert {d_i^s-\hat{d_i^s}}\rvert<C_1} + max(c,0)_{\tau\lvert {d_i^s-\hat{d_i^s}}\rvert>C_2}
\end{align}
where $\tau$ is an indicator function which equals 1 when the condition is satisfied and 0 otherwise. The function aims at increasing the confidence if the distance between predicted disparity and ground truth is smaller that a threshold $C_1$ and decreasing the confidence if the distance is larger than a threshold $C_2$. In our work, we set $C_1=0.5$ and $C_2=1.5$.  

The final loss function is a combination of losses for the disparity, surface normal, and confidence for all pixels across 4 scales:
\begin{align}
 L^{total} = \sum_{s=0}^3
 \lambda_s(\lambda_{d}L_s^d+\lambda_{n}L_s^n +\lambda_{c}L_s^c)
\end{align}
where $\lambda_s$ is a scalar for adjusting the loss weight at scale $s$. $\lambda_d$, $\lambda_n$ and $\lambda_c$ are coefficients for disparity, surface normal and confidence losses, respectively. More details are provided in Section \ref{subsec:implementation}. 

\section{Experiment}\label{sec:exp}
In this section, we will first introduce the training datasets as well as the detailed training scheme. Then we provide a thorough quantitative analysis of our proposed model on the corresponding datasets.

\subsection{Datasets and Evaluation Metrics}
We validate the effectiveness on three public datasets which are Scene Flow \cite{dispnetc}, KITTI 2015 \cite{Kitti2015}, and Middlebury \cite{middlebury2014}. The Scene Flow dataset consists of 39,824 pairs of synthetic stereo RGB images with a full resolution of 960$\times$540 among which 35,454 pairs are split as the training set and the rest 4,370 pairs are split as the testing set. To facilitate the research of autonomous driving, the KITTI 2015 collects 400 pairs of real scenes with sparse depth ground truth annotated by LiDAR, where 200 pairs are for training and the other 200 pairs are utilized for testing. Only 33 pairs of stereo images are available for the MiddleBurry2014 dataset so we conduct a generalization evaluation on it. The standard end-point error (EPE) is reported on the Scene Flow dataset and we follow the standard evaluation protocol to submit our predicted results to the KITTI Benchmark from where the D1 error is obtained. 

\subsection{Normal Estimation Supervision} \label{subsec:nes}
For the Scene Flow dataset, the provided disparity labels are pixel-wise annotated which makes it possible to directly introduce the Sobel kernel for calculating gradient along $x$ and $y$ axis, respectively. The primary challenge comes from the KITTI datasets because of the sparse disparity annotations. To tackle this problem, we first leverage the state-of-the-art model LEAStereo \cite{leastereo} to generate the dense pseudo disparity labels following the Sobel kernel to calculate the pseud normal ground truth $N_{pseudo}$. Then we search for the local region where the disparity ground truth is available for the neighboring 8 pixels to obtain the normal ground truth $N_{sparse}$ as well as the ground truth position mask $M_{sparse}$. The final normal ground truth $N_{gt}$ is then calculated as follows:
\begin{align}
    N_{gt} = N_{pseudo} * (1-M_{sparse}) + N_{sparse} * M_{sparse}
\end{align}

\begin{figure}
	\centering
	\includegraphics[width=0.92\linewidth]{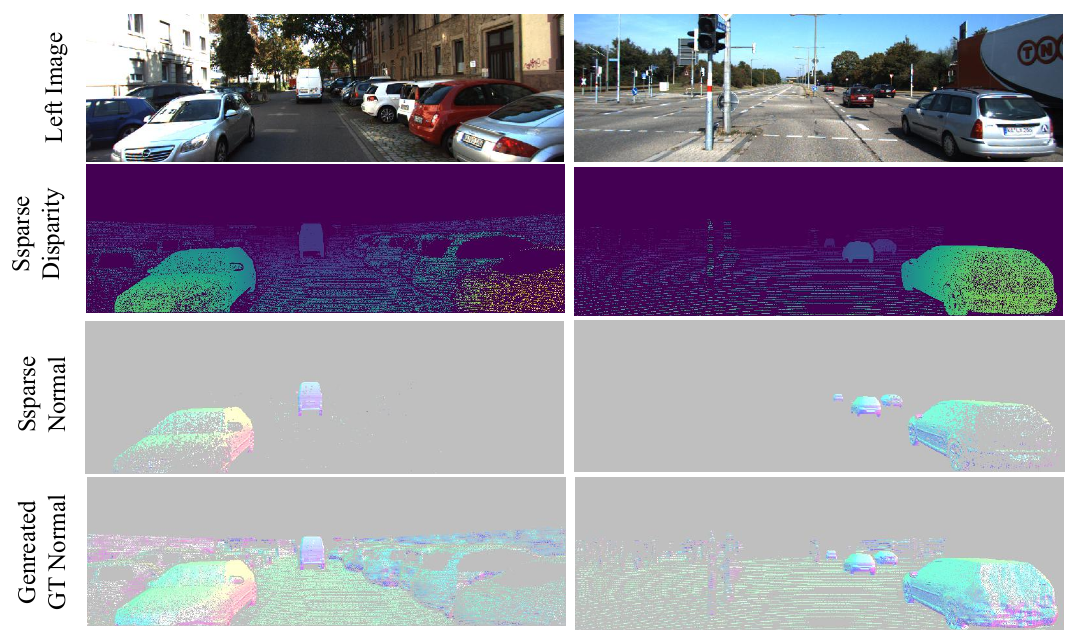}
	\caption{Visualization of our generated surface normal ground truth on KITTI 2015. Row one is the left image, row two is the gt sparse disparity, row three is the sparse surface normal generated with neighbour-existing disparities.The last row is our generated disparity with the assistance of pseudo disparity. }
	\label{fig:kitti_normal}
	\vspace{-0.8 em}
\end{figure}

To fully exploit the limited sparse normal information and alleviate the influence of pseudo disparity of low quality, we further calculate the end-point error between the pseudo disparity labels and the sparse disparity ground truth to obtain an index $E_{index}$ where the EPE is greater than 1 disparity. The loss of normal estimation $L_N$ is computed as follows:
\begin{align}
    L_N = &S(N_{gt}, \hat{N})* M_{sparse} * 1.5 \\ \nonumber
          &+ S(N_{gt}, \hat{N}) * (1-M_{sparse}) * E_{index} * 1.0 \\ \nonumber
          &+ S(N_{gt}, \hat{N}) * (1-M_{sparse}) * (1-E_{index}) * 0.5,
\end{align}
where $S(x,y)$ denotes the SmoothL1 loss function mentioned before and $\hat{N}$ represents the predicted normal. Our generated surface normal ground truth is shown in Figure \ref{fig:kitti_normal}

\subsection{Implementation Details} \label{subsec:implementation}
We implemented our NINet by PyTorch\cite{pytorch} and trained our model with Adam(momentum=0.9, beta=0.999) as the optimizer. We randomly cropped the input to 320$\times$640 on the KITTI 2015 $\&$ 2012 and the Scene Flow datasets. The whole training process is performed on 2 NVIDIA RTX 2080ti GPUs with a batch size of 8. The max disparity search range is set as 192 pixels. The coefficients of four disparity losses at each scale are correspondingly set as $\lambda_0=1.0$, $\lambda_1=\lambda_2=0.8$, and $\lambda_3=0.6$. The weights for disparity, surface normal and confidence are set as $\lambda_d =5$, $\lambda_n =50$, $\lambda_n =1$.

For the Scene Flow dataset, we trained the networks for 150 epochs in total with an initial learning rate of 1e-3. After the first 20 epochs of training, we decrease our learning rate to 4e-4,1e-4,4e-5 at the epoch of 20, 40, 100. Note that to enhance the reliability of spatial propagation guided by normal, the normal estimation branch is trained independently for 40 epochs before the joint learning of disparity and normal.

For the KITTI 2015 and the KITTI 2012 datasets, we trained our networks based on the model pre-trained on the Scene Flow dataset. We first mixed these two datasets with $300$ pairs of images in total for the first $200$ epochs of training to obtain the model with the best performance on the validation set. Another $200$ epochs of fine-tuning are employed on the KITTI 2015 and the KITTI 2012 datasets respectively. The learning rate is constantly set as 1e-5 for the whole training process. 

\begin{table*}
    \setlength{\tabcolsep}{2.6mm}
    \centering
    \caption{Ablation study of our proposed method on the Scene Flow dataset. The best performance is stressed in bold. Our proposed affinity-aware residual learning and non-local spatial propagation modules significantly improve the disparity estimation. The best results are stressed in bold. Our proposed modules efficiently improve the stereo matching performance.}
    \begin{tabular}{c|c|c|c|c|c|c|c|c} \hline
         \multirow{2}{*}{Method} & \multirow{2}{*}{NR Branch} & \multicolumn{2}{c|}{Affinity-aware Residual} & \multicolumn{2}{c|}{Spatial Propagation} & \multirow{2}{*}{EPE} & \multirow{2}{*}{P1 Error} & \multirow{2}{*}{P3 Error} \\ \cline{3-6}
         & &  attention & dynamic filtering & local & non-local & &  \\ \hline
         Baseline & & & & & & 0.96 & 12.65$\%$ & 5.94$\%$ \\ \hline
         Baseline-R & \cmark & & & & & 0.73 & 9.58$\%$ & 4.72$\%$ \\ \hline
         Baseline-RD & \cmark & & \cmark & & & 0.66 & 8.06$\%$ & 4.42$\%$ \\ \hline
         Baseline-RA & \cmark & \cmark  & & & & 0.63 & 7.90$\%$ & 4.18$\%$ \\ \hline
         Baseline-ARL & \cmark & \cmark & \cmark & & & 0.60 & 7.42$\%$ & 4.00$\%$ \\ \hline
         Baseline-LDP & \cmark & & & \cmark & & 0.65 & 8.02$\%$ & 4.20$\%$ \\ \hline
         Baseline-NDP & \cmark & & & & \cmark & 0.60 & 7.39$\%$ & 3.98$\%$ \\ \hline
         Baseline-ARL-LDP & \cmark & \cmark & \cmark & \cmark & & 0.54 & 6.65$\%$ & 3.71$\%$ \\ \hline
         Baseline-ARL-NDP & \cmark & \cmark & \cmark & & \cmark & \textbf{0.51} & \textbf{6.17}$\%$ & \textbf{3.53$\%$} \\ \hline
    \end{tabular}
    \label{tab:ablation}
\end{table*}

\begin{table*}
    \setlength{\tabcolsep}{1.2mm}
    \centering
    \caption{Comparison of our proposed method and other remarkable works on the Scene Flow dataset. Among all the competing methods, our NINet ranks the 3rd for end-point error. The best performance is emphasized in bold.}
    \begin{tabular}{c|c|c|c|c|c|c|c|c|c} \hline
         Model & DispNetC\cite{dispnetc} & PSMNet\cite{psmnet} & GwcNet\cite{gwcnet} & EDNet\cite{EDNet} & HITNet\cite{HitNet} & LeaStereo\cite{leastereo} & CSPN\cite{CSPN} & ACVNet\cite{acvnet} & NINet (Ours) \\ \hline
         EPE & 1.68 & 1.09 & 0.76 & 0.63 & \textbf{0.41} & 0.78 & 0.78 & 0.48 & 0.51 \\ \hline
    \end{tabular}
    \label{tab:SF}
\end{table*}

\begin{table}
    \setlength{\tabcolsep}{2.6mm}
	\centering
	\caption{Benchmark results on KITTI 2015 test sets. “Noc” and “All” indicate the percentage of outliers averaged over ground truth pixels of non-occluded and all regions, respectively. “fg” and “all” indicate the percentage of outliers averaged over the foreground and all ground truth pixels, respectively. Our NINet achieves the best performance on the foreground pixels with a competitive inference speed.}
	\begin{tabular}{c|cc|cc|c} \hline
		\multirow{2}{*}{Method} & \multicolumn{2}{c|}{Noc (\%)} & \multicolumn{2}{c|}{All (\%)} & \multirow{2}{*}{Time (s)} \\ \cline{2-5}
		& fg & all & fg & all & \\ \hline
		GANet \cite{ganet} & 3.37 & 1.73 & 3.82 & 1.93 & 0.36 \\ 
		GCNet \cite{gcnet} & 5.58 & 2.61 & 6.16 & 2.87 & 0.9 \\ 
		PSMNet \cite{psmnet} & 4.31 & 2.14 & 4.62 & 2.32 & 0.41 \\ 
		GwcNet \cite{gwcnet} & 3.49 & 1.92 & 3.93 & 2.11 & 0.32 \\ 
		SegStereo \cite{segstereo} & 3.70 & 2.08 & 4.07 & 2.25 & 0.6 \\
		HD$^3$ \cite{hd3} & 3.43 & 1.87 & 3.63 & 2.02 & 0.14 \\
		Bi3D \cite{Bi3D} & 3.11 & 1.79 & 3.48 & 1.95 & 0.48 \\
		AANet \cite{AANET} & 4.93 & 2.32 & 5.39 & 2.55 & 0.075 \\
		DispNetC \cite{dispnetc} & 3.72 & 4.05 & 4.41 & 4.34 & 0.06 \\
		FADNet \cite{fadnet} & 3.07 & 2.59 & 3.50 & 2.82 & 0.05 \\
		EDNet\cite{EDNet}  & 3.33 & 2.31 & 3.88 & 2.53 & 0.05 \\ 
		ACVNet\cite{acvnet} & 2.84 & 1.52 & 3.07 & \textbf{1.65} & 0.20 \\
		CSPN\cite{CSPN} & 2.67 & 1.61 & 2.88 & 1.74 & 1.0\\
		HITNet \cite{HitNet} & 2.72 & 1.74 & 3.20 & 1.98 & 0.02 \\
		LEAStereo\cite{leastereo} & 2.65 & \textbf{1.51} & 2.91 & \textbf{1.65} & 0.30 \\ \hline
		NINet (Ours) & \textbf{2.53} & 1.83 & \textbf{2.80} & 1.98 & 0.17 \\ \hline
	\end{tabular}
	\vspace{-1em}
	\label{tab:kitti2015}
\end{table}
\begin{figure}
	\centering
	\includegraphics[width=0.94\linewidth]{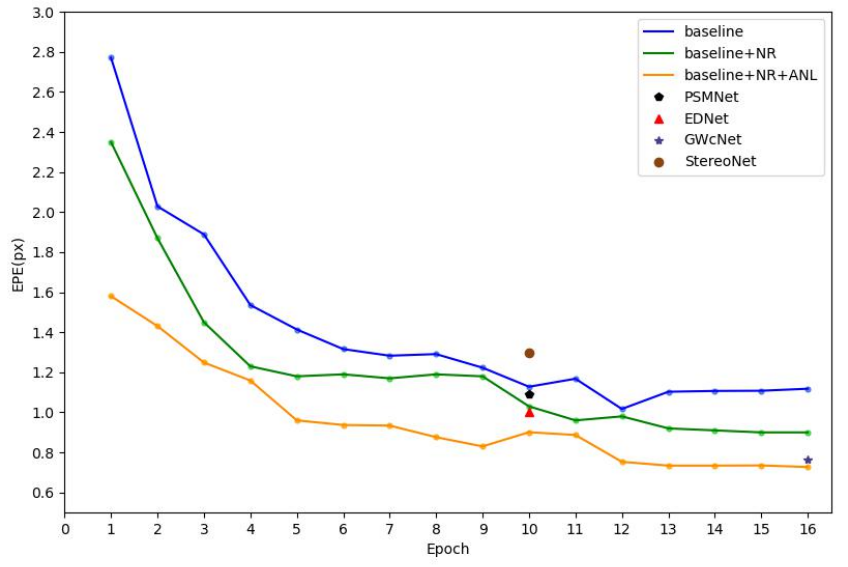}
	\caption{Comparison among our proposed methods with different settings on the Scene Flow testing set as well as other SOTA works. It is obvious that our proposed modules significantly facilitate the speed of convergence.}
	\label{fig:convergence}
	\vspace{-0.8em}
\end{figure}

\begin{figure}
	\centering
	\includegraphics[width=0.98\linewidth ]{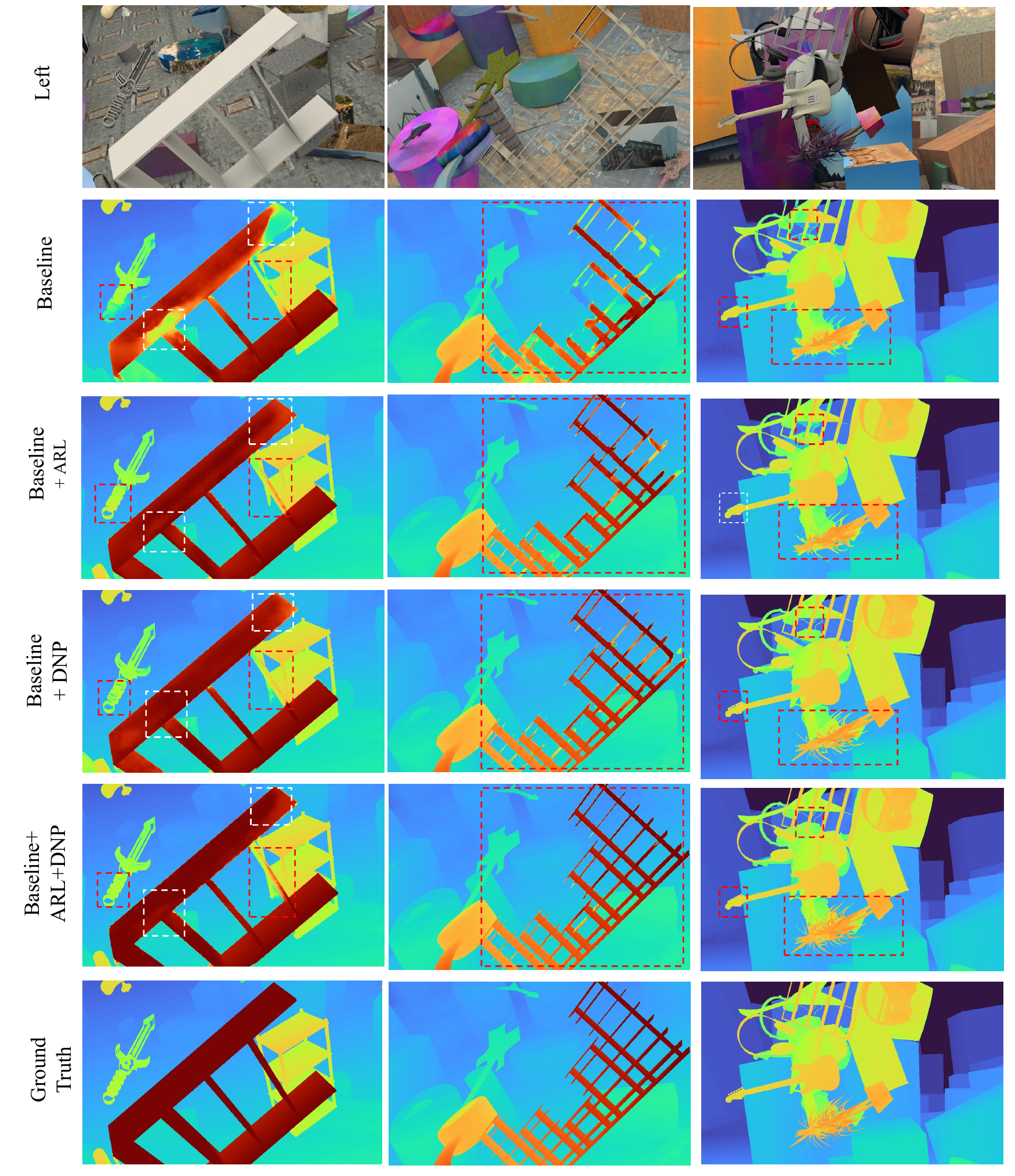}
	\caption{Visualization of the ablation study on Scene Flow dataset. With the assistance of our proposed modules, thin structures and more consistent disparity can be preserved.}
	\label{fig:SF_ablation}
	\vspace{-1em}
\end{figure}

\subsection{Ablation Study}
Based on the performance shown in Table \ref{tab:ablation}, it can be directly analyzed that our proposed non-local disparity propagation and affinity-aware residual learning modules effectively improve the performance. The joint learning scheme with surface normal improves the baseline with an obvious margin. The employment of the affinity-aware residual learning module which combines both attention and dynamic filtering operations significantly decreases the EPE to 0.60. The disparity level non-local propagation further improves the result as reflected by the EPE of baseline-ARL-NDP, which is 0.51.
The second column provides straightforward evidence that our NINet successfully preserves the detailed outline of the shelf with consistent disparity while other settings fail to reach this result. To further provide a more comprehensive interpretation of the superiority of our method, we visualize the converging speed of our methods as well as other competing approaches as shown in Figure \ref{fig:convergence}. With the assistance of the proposed modules, our method not only achieves the lower initial EPE but outperforms other works at the corresponding epochs as well. Note that, the competitive result of GwcNet is mainly attributed to the application of 3D convolution which is discarded in our work.

\begin{table}
    \setlength{\tabcolsep}{4mm}
    \centering
    \caption{Evaluation for bordered and occluded regions on the Scene Flow dataset with our proposed modules.}
    \begin{tabular}{c|c|c} \hline
         Method & Bordered EPE & Occluded EPE \\ \hline
         Baseline & 1.95 & 2.31 \\ \hline
         +ARL & 1.63 & 1.92 \\ \hline
         +NDP & 1.44 & 1.81 \\ \hline
         +NDP+ARL & 1.31 & 1.71 \\ \hline
    \end{tabular}
    \label{tab:bordered_error}
    \vspace{-1em}
\end{table}

\begin{table}
    \setlength{\tabcolsep}{1.3mm}
    \centering
    \caption{Generalization evaluation on the MiddleBurry 2014 dataset. Our NINet achieves the best generalizability.}
    \begin{tabular}{c|c|c|c|c|c|c|c|c} \hline
         \multirow{2}{*}{Method} & \multicolumn{2}{c|}{Bad2.0} & \multicolumn{2}{c|}{Bad4.0} & \multicolumn{2}{c|}{RMSE} & \multicolumn{2}{c}{AvgErr} \\ \cline{2-9}
         & Occ & Non & Occ & Non & Occ & Non & Occ & Non \\ \hline
         AANet & 38.6 & 42.5 & 25.3 & 28.9 & 23.3 & 21.4 & 12.4 & 10.7 \\ \hline
         EDNet & 69.5 & 66.8 & 57.1 & 54.0 & 21.9 & 20.4 & 44.2 & 41.2 \\ \hline
         GwcNet & 34.9 & 30.8 & 22.7 & 19.4 & 27.0 & 25.4 & 13.4 & 11.8 \\ \hline
         FADNet & 31.3 & 27.0 & 20.9 & 17.1 & 21.8 & 19.6 & 11.2 & 9.5\\ \hline
         NINet(Ours) & \textbf{24.5} & \textbf{20.2} & \textbf{15.8} & \textbf{12.5} & \textbf{18.8} & \textbf{18.3} & \textbf{8.0} & \textbf{7.8} \\ \hline
    \end{tabular}
    \label{tab:generalization}
    \vspace{-1em}
\end{table}


\subsection{Performance Evalution}
In this subsection, we will mainly focus on the comparison between our method and other top-performing methods.

\begin{figure}
	\centering
	\includegraphics[width=0.93\linewidth]{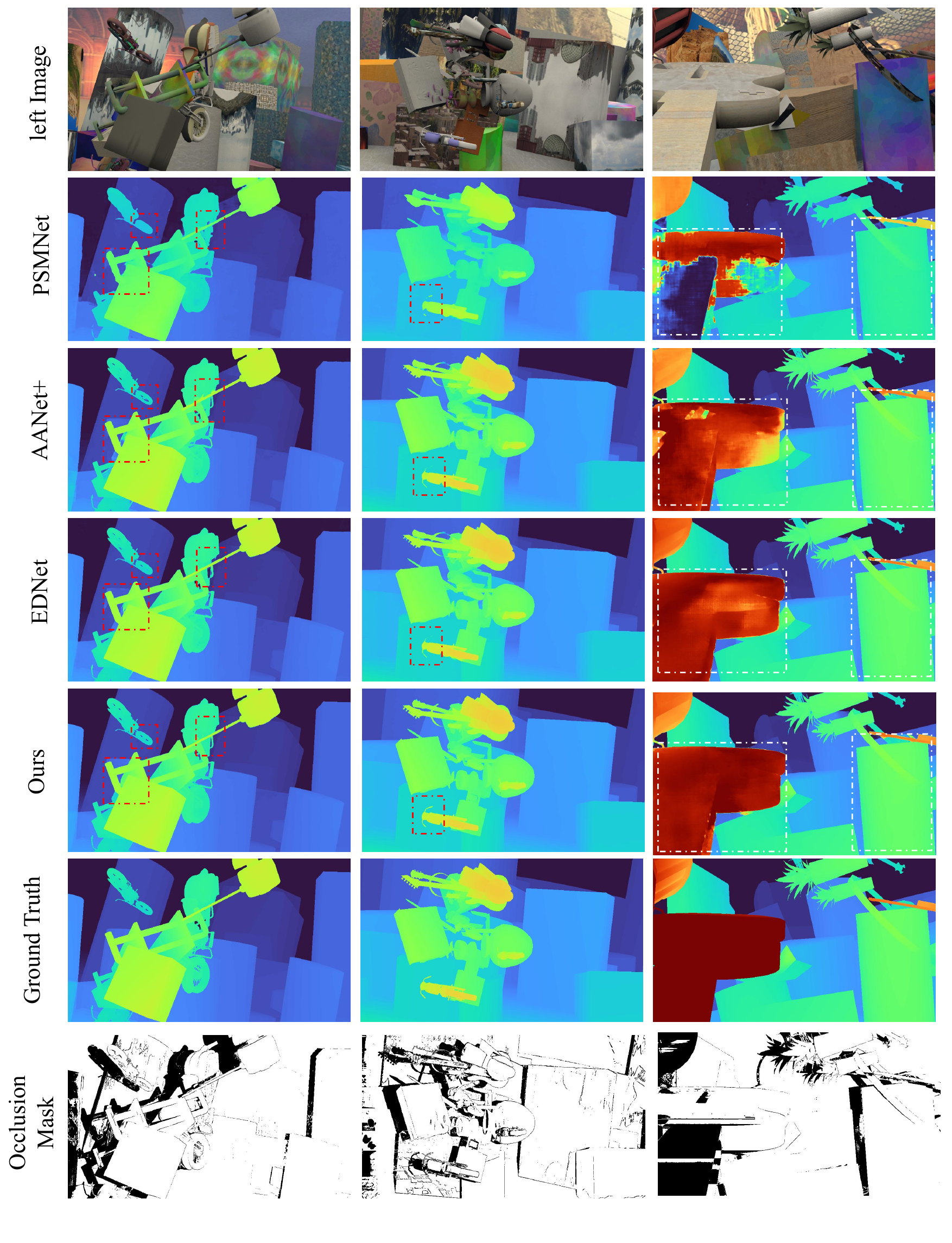}
	\caption{SceneFlow performance compares with other superior works.}
	\label{fig:SF_compare}
\end{figure}

\textbf{Scene Flow} As shown in Table \ref{tab:SF}, our method outperforms most of the superior approaches including LEAStereo \cite{leastereo} and ranks 3rd among all the competing works. As illustrated in Figure \ref{fig:SF_compare}, our method produces sharper and more detailed disparity estimation. A representative instance comes from the third column that our NINet achieves consistent disparity estimation for the plank while results from other methods are less convincing. Table \ref{tab:bordered_error} specifically reveals the effectiveness of proposed modules in both bordered and occluded regions.

\textbf{KITTI 2015} Table \ref{tab:kitti2015} demonstrates the performance of several remarkable methods in the KITTI 2015 leaderboard. When evaluating disparity estimation in foreground areas, our approach even outperforms the HITNet by 6.99$\%$ in non-occluded regions and ranks 1st among all the published works, which strongly validates the effectiveness of our model.

\textbf{MiddleBurry 2014} Generalizability is also a necessary indicator when comprehensively evaluating learning-based stereo matching methods and thus we conduct a generalization experiment on MiddleBurry 2014 dataset. As shown in Table \ref{tab:generalization}, our proposed method obtains the best performance when transferring to a new data domain without finetuning. Compared with 3D convolution-based GwcNet, our method decreases the average error from 13.4 to 8.0 and 11.8 to 7.8 in occluded regions and non-occluded, respectively.


\begin{figure}
	\centering
	\includegraphics[width=1.03\linewidth]{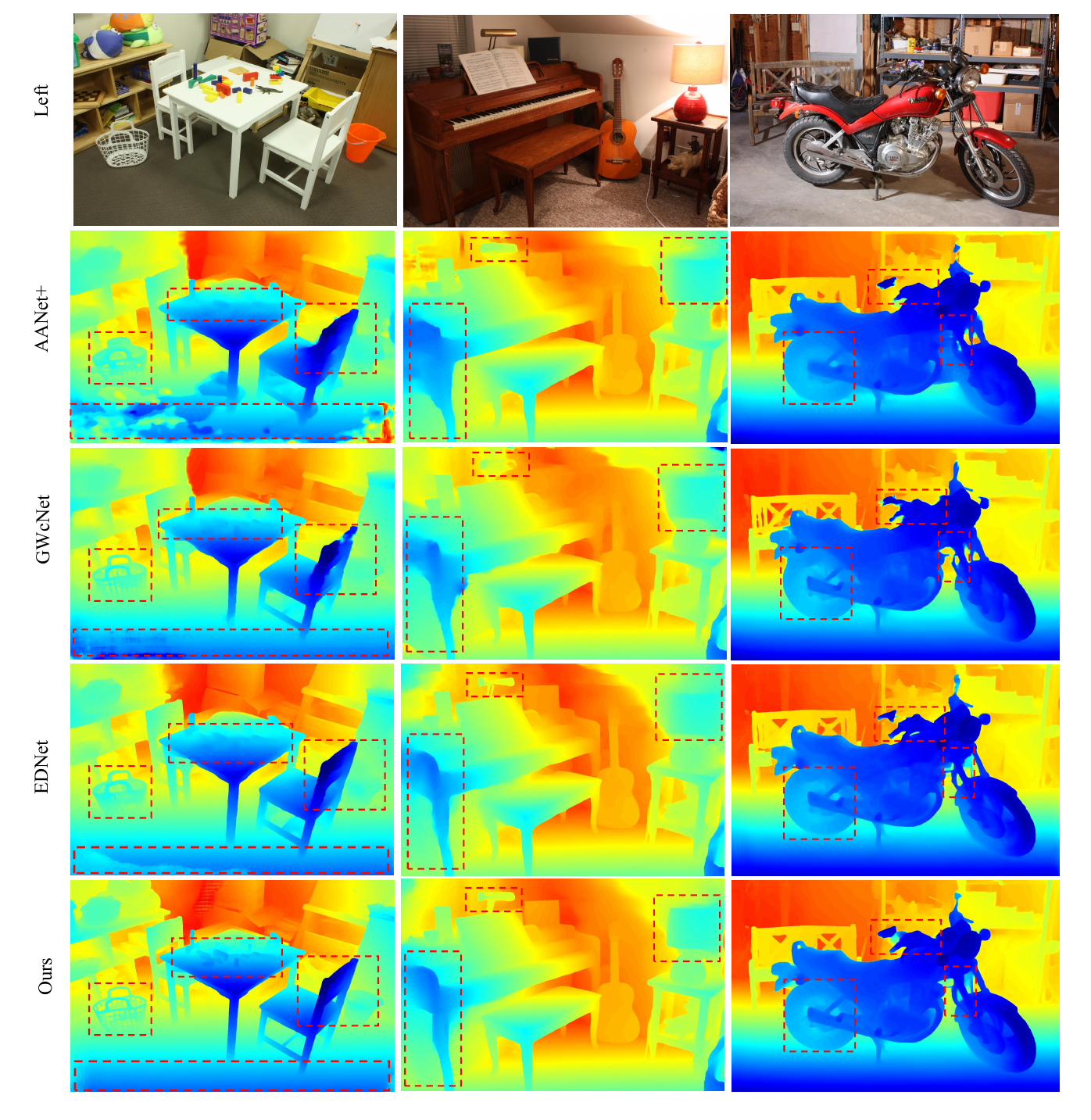}
	\caption{Generalization ability on MiddlyBurry 2014 dataset.}
	\label{fig:MB}
\end{figure}
\section{Conclusion}\label{sec:conclusion}
In this paper, we propose a normal incorporated joint learning framework named NINet to explicitly leverage the plane information for intuitive geometric guidance. A non-local spatial propagation (NDP) module is also introduced to dynamically select points with a similar affinity for further propagation at the disparity level. An affinity-aware residual learning (ARL) module is then designed to impose consistency within the same plane according to a local affinity matrix. Extensive experiments on several datasets have demonstrated the effectiveness as well as the generalization ability of our proposed mothed. By the time we finished it, our method ranked 1st on the KITTI 2015 dataset for foreground objects.

\bibliographystyle{ACM-Reference-Format}
\balance
\bibliography{acmart}

\clearpage
\appendix

\section{Normal Estimation Sub-Net}\label{sec:nomal_subnet}

\begin{figure}[!htbp]
    \centering
    \includegraphics[width=0.95\linewidth]{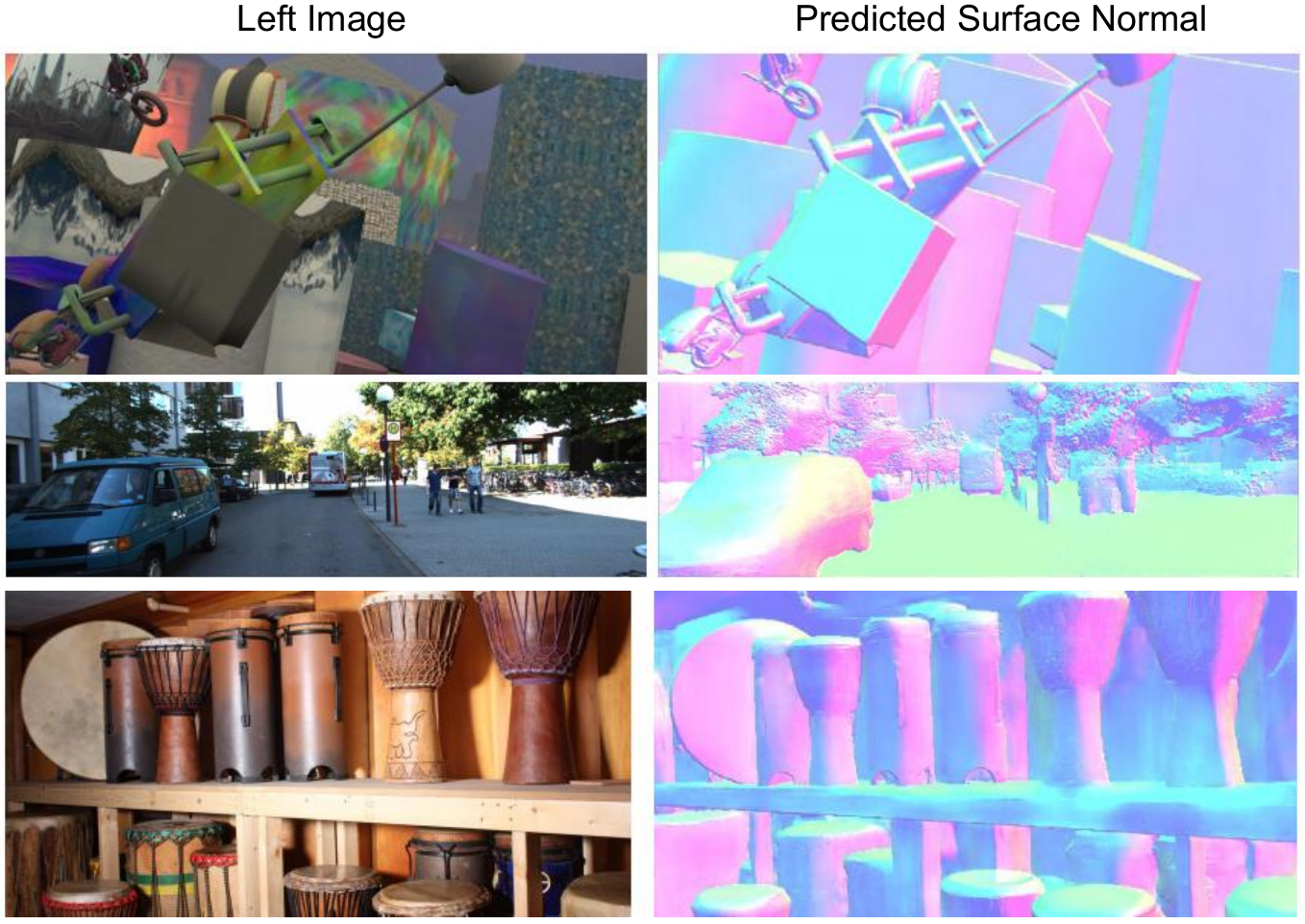}
    \caption{Normal Estimation Result on SceneFlow,KITTI2015 and MiddleBurry testing set.}
    \label{fig:normal_SF}
\end{figure}

Our Normal Estimation Sub-Net is illustrated in Figure \ref{fig:normal_subnet}. We directly use the shared encoder part to provide the feature and adopt the network structure of DispNetC[1] to estimate the surface normal in a coarse-to-fine manner. For residual learning(Residual Blocks in blue ), we use a shallow U-Net architecture to take features at the current scale together with upsampled predicted surface normal at the previous scale and left-right image pairs as input to predict the surface normal residual at the current scale.
Besides, we normalized the output of each scale directly on the feature channels(3), making it more in line with the definition of the surface normal. Figure \ref{fig:normal_SF} illustrate visualization results on SceneFlow and KITTI 2015 and MiddleBurry 2014 datasets.

\section{Non-Local Disparity Propagation at different scales.}

In this section, by visualizing the disparities estimation results at different resolutions, we show how our proposed NDP module improves the quality of the predicted disparity through spatial propagation, especially in thin structures, edges, and occluded regions. As is shown in Figure \ref{fig:disparity_refinement}, the non-local disparity propagation based on surface normal and other semi-context information performs well at thin structures and edges especially at low scales to alleviate the disparity discontinuity and blurring issues. After spatial propagation, our disparity preserves more structural details, which facilitates more accurate depth results for downstream tasks and thus constructs more accurate AR/VR applications.


\section{Spatial attention and affinities distribution in ARL module}

As mentioned in the main paper, we have proposed an ARL module that manages to optimize disparity estimation results at the feature level. There are two main tricks in the ARL module for efficient residual learning. One is to apply the dynamic spatial attention mechanism to generate attention maps at different refinement scales. The attention maps imply where should be highlighted as well as where should be ignored. It promotes the network to efficiently optimize the disparity estimation results for different locations at different scales.

\begin{figure}[!htbp]
    \centering
    \includegraphics[width=1.0\linewidth]{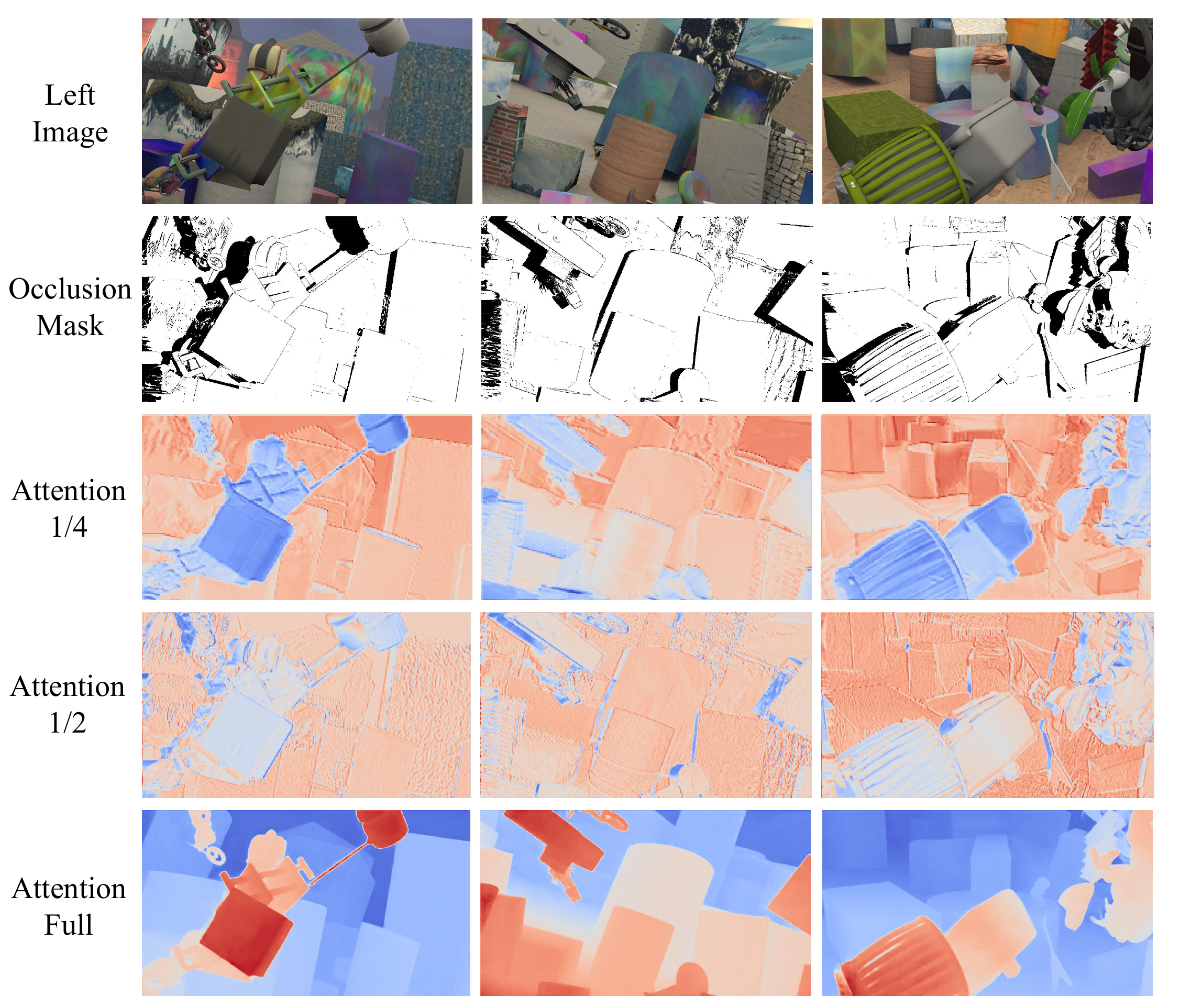}
    \caption{Attention Maps at different scales(Range from 1/4~full scale). Different scales pay different attention to different regions, such as foreground, texture-less planes, occluded areas and background.}
    \label{fig:attention}
\end{figure}

\begin{figure*}[htbp]
    \centering
    \includegraphics[width=0.85\linewidth]{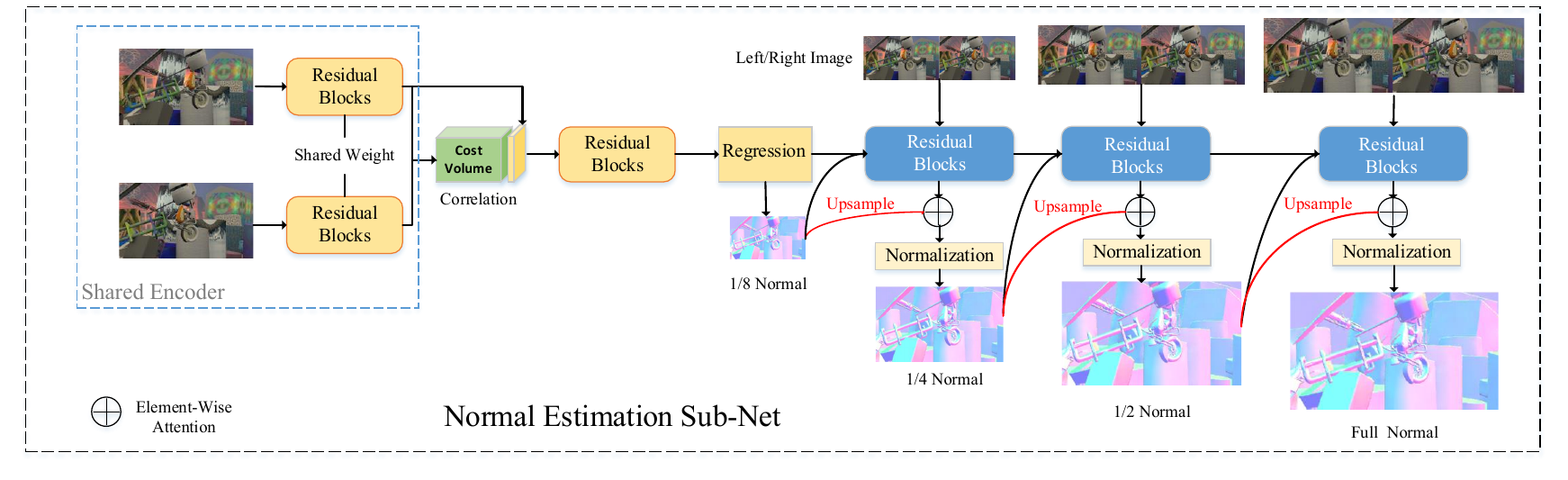}
    \caption{Normal Estimation Sub-Net Architecture. The feature encoder is shared with the disparity estimation branch. The whole Sub-Net adopts the DispNetC \cite{dispnetc} architecture with a residual learning module. We normalized the predicted disparity to satisfy the geometry constrain.}
    \label{fig:normal_subnet}
\end{figure*}

\begin{figure*}
    \centering
    \includegraphics[width=0.7\linewidth]{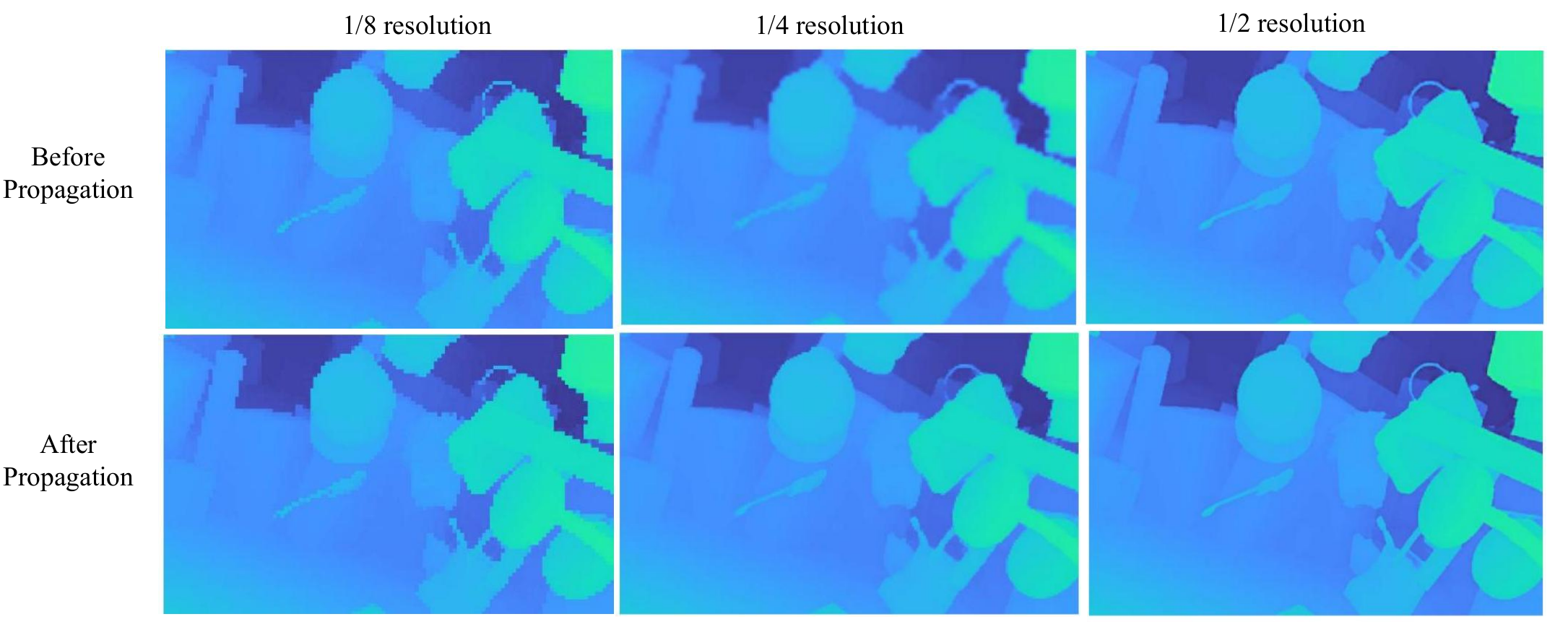}
    \caption{Disparity refinement at different scales. It clearly shows that our proposed non-local propagation witness a clear edges and structures at different disparity scales, which alleviates the blurring and breakage issues at the edges of the image.}
    \label{fig:disparity_refinement}
\end{figure*}

As shown in Figure \ref{fig:attention}, the dynamic spatial attention mechanism tends to assist residual learning at different levels and focus on regions with different attributes. The attention distribution at the 1/4 scale mainly lies in the background objects, while the distribution of attention at 1/2 resolution is highly consistent with that of the occlusion mask. This enables the ARL module to be occlusion-aware when refining the disparity. Besides, the attention map at full scale shows bigger attention to foreground object and planes that is texture-less, which benefit the refinement in such regions.

\begin{figure*}
    \centering
    \includegraphics[width=0.6\linewidth]{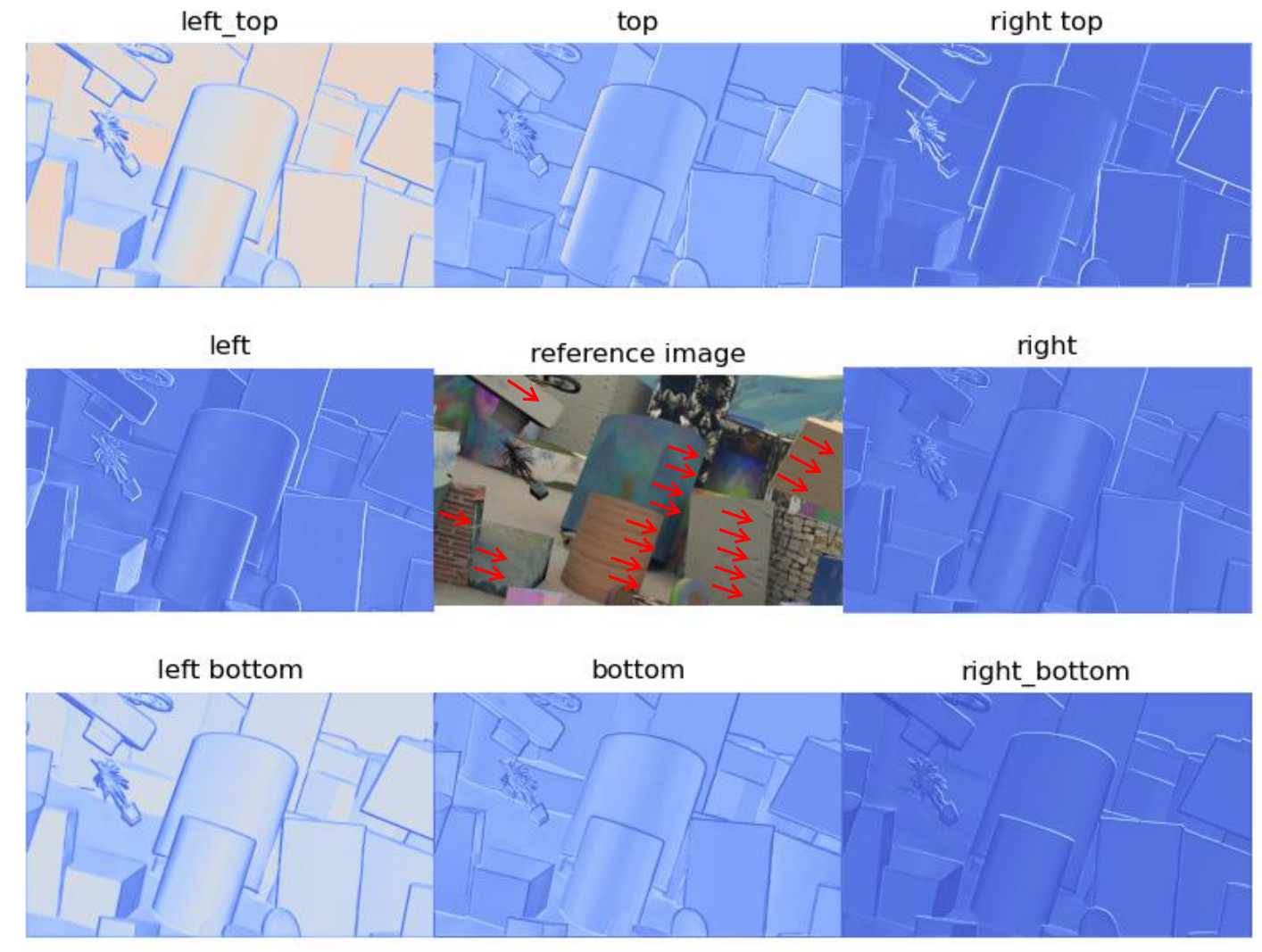} 
    \caption{A sample of affinity distribution of local 8 neighbors in ARL module.}
    \label{fig:filtering}
\end{figure*}

The other trick of the ARL module is the local affinity filtering module whose intermediate affinity visualization result of the local 8 neighbors can be seen in Figure \ref{fig:filtering}. As the figure illustrated, the red arrow in reference images shows the plane direction, as we can see that the affinity value at the left-top value is bigger than the other direction, which indicates the direction of the feature aggregation should be. By applying the local affinity filtering module, our ARL module can make full use of surface normal information for better feature aggregation.

\end{document}